%% file: main.tex
\def\cleanversion{}
\documentclass[journal]{IEEEtran}

\usepackage{amsmath,amssymb,amsfonts}
\usepackage{graphicx}
\graphicspath{{figures/}}
\usepackage{cite}
\usepackage{textcomp}
\usepackage{booktabs}
\usepackage{multirow}
\usepackage{bm}
\usepackage{array}
\input{revmarks}

\ifdefined\cleanversion\newcommand{\revcell}{\color{black}}\else\newcommand{\revcell}{\color{\revcolor}}\fi
\newcolumntype{C}{>{\revcell}c}\newcolumntype{L}{>{\revcell}l}
\usepackage[hidelinks]{hyperref}

\makeatletter
\newcommand{\markrevisedfloat}{
  \let\originalmakecaption\@makecaption
  \long\def\@makecaption##1##2{{\revcell\originalmakecaption{##1}{##2}}}
}
\makeatother

\begin{document}

\title{Robust Well-Log Reconstruction Across Point Gaps and Whole Wells via Boundary-Gated Soft-Thresholding Kalman Fusion}

\author{Yongan~Zhang,
        Zhihan~Qiu,
        Yizheng~Li,
        and~Youzhuang~Sun
\thanks{This work was supported by the China National Petroleum Corporation under Grant ZD2019-183004 and the Fundamental Research Funds for the Central Universities under Grant 20CX05019A. (\textit{Corresponding author: Youzhuang Sun.})}
\thanks{Yongan Zhang and Yizheng Li are with the State Key Laboratory of Climate Resilience for Coastal Cities, Department of Building Environment and Energy Engineering, The Hong Kong Polytechnic University, Hong Kong SAR, China (e-mail: a374456248@outlook.com; yzli@eitech.edu.cn).}
\thanks{Zhihan Qiu is with Shandong University of Science and Technology, Qingdao, China (e-mail: 3163003266@qq.com).}
\thanks{Youzhuang Sun is with the School of Computer Science and Technology, China University of Petroleum (East China), Qingdao 266580, China (e-mail: 905736579@qq.com).}}

\maketitle

\begin{abstract}\rev{Subsurface characterization for hydrocarbon exploration and geological CO\(_2\) storage relies on wireline logs to calibrate seismic images, yet washouts, tool failures and partial runs leave multiscale gaps. However, existing methods ignore borehole spike contamination, measured \mbox{gap-edge} constraints and \mbox{cross-basin} petrophysical shifts, causing deep networks to suffer from boundary steps on short gaps and domain degradation across basins while interpolation and Gaussian smoothers fail across \mbox{multi-bed} intervals or track outliers. To overcome these failures, this study proposes \mbox{boundary-gated} \mbox{soft-thresholding} Kalman fusion (\mbox{ST-KF}), a \mbox{measurement-guided} framework unifying multiresolution wavelet shrinkage, \mbox{distance-gated} residual prior learning and \mbox{closed-form} robust Kalman smoothing under learning--inference decoupling. Across two public benchmarks totaling 263 wells in the Norwegian North Sea and New Zealand's Taranaki Basin, \mbox{ST-KF} ranks first over 0.15--100-m anchored gaps with 0.09\% parameter overhead, lowering pooled RMSE by 5.0\%--8.3\% for bulk density and 6.4\%--15.6\% for compressional slowness against the strongest of nine baselines and by 0.7\%--4.1\% for neutron porosity and 1.3\%--2.9\% for photoelectric factor against the four strongest baselines. Ablation and sensitivity analyses confirm operator synergy and flat stability, where removing \mbox{state-space} fusion increases \mbox{short-gap} errors by up to 58.8\%, while \mbox{zero-shot} transfer cuts anchored errors by 14\%--68\%. This study reveals that local borehole continuity transfers across basins whereas \mbox{cross-property} mappings shift with regional geology, establishing a \mbox{measurement-guided} paradigm for subsurface log reconstruction.}\end{abstract}


\begin{revblock}

\section{Introduction}

\label{h-s1}

Geological knowledge of sedimentary basins, from hydrocarbon exploration to CO\(_2\) storage and \mbox{pore-pressure} prediction, depends on measurements of rock properties at depth. Reflection seismics images entire basins but resolves layers only at the decametre scale. As a tool string comprising multiple sensors moves along each borehole, wireline logs measure formation properties in place every 0.15 m and provide the ground truth that calibrates seismic images \cite{ellis2007well}. Bulk density (RHOB) and compressional slowness (DTC) are fundamental inputs that together define acoustic impedance, which ties wells to seismic data \cite{zhang2020seismic}. The same two logs enter porosity estimation \cite{mehrabi2026dual}, \mbox{pore-pressure} prediction \cite{tassone2019geological}, geomechanical models \cite{chen2020physics,maalouf2025geomechanical} and the monitoring of injected CO\(_2\) \cite{naseer2024evaluation}. In field operations, the completeness of \mbox{well-log} datasets, particularly RHOB and DTC, is affected by several factors. Density pads lose contact in \mbox{washed-out} intervals, sonic tools \mbox{cycle-skip} in \mbox{gas-bearing} or fractured zones, older wells were run with partial tool strings, and some wells were never logged for the target property at all \cite{akinnikawe2018synthetic,gama2025imputation}. Laboratory measurements in carbonate rocks also show that brine saturation increases acoustic attenuation, with contributions from \mbox{pore-fluid} flow and scattering \cite{bouchaala2024attenuation}. Missing intervals therefore range from isolated corrupted samples to tens of metres and to entire wells. Within the \mbox{reservoir-evaluation} workflow, the quality of \mbox{well-log} reconstruction affects not only the accuracy of individual parameter estimates but also downstream fluid identification and reserve calculations.

Early approaches to \mbox{well-log} reconstruction relied primarily on empirical formulas, multivariate regression, and spatial interpolation. Gardner et al.~(1974) related density to velocity with an empirical power law \cite{gardner1974formation}, which holds only within a given lithology. \mbox{Machine-learning} regression then predicted the missing log from the remaining logs at the same depth, for example with gradient boosting in Akinnikawe et al.~(2018) \cite{akinnikawe2018synthetic} and with random forests that also estimate uncertainty in Feng et al.~(2021) \cite{feng2021rf}. Sequence models added depth context. Zhang et al.~(2018) generated missing logs with recurrent networks \cite{zhang2018lstm}, Zhang et al.~(2022) predicted compressional and shear slowness with a convolutional recurrent network \cite{zhang2022convolutional}, and Lin et al.~(2023) reconstructed logs lost to borehole conditions with sequence \mbox{self-attention} \cite{lin2023missing}. Domain knowledge entered through \mbox{rock-physics} guidance of \mbox{shear-sonic} prediction in Zhao et al.~(2023) \cite{zhao2023rock} and through \mbox{geology-constrained} adversarial training in Guo et al.~(2026) \cite{guo2026gan}. The most recent \mbox{gap-filling} models also take the observed part of the target as input, as in the conditional diffusion model of Meng et al.~(2024), which adds a measure of uncertainty \cite{meng2024missing}. Gama et al.~(2025) compared such networks with interpolation on three public collections and found that the networks win on long gaps but that their ranking changes with gap length and dataset \cite{gama2025imputation}. In all of these models the measured samples next to a gap are soft inputs rather than values that the output must reproduce at the gap edge, and no single model has been shown to hold across gap regimes.

These routes leave three coupled failure modes across measurement corruption, representation learning and posterior fusion. At the measurement level, washouts and sonic cycle skipping inject isolated target spikes while predictor logs carry tool noise \cite{ellis2007well}, and detached preprocessing \cite{honorio2012well} passes surviving spikes downstream as ground truth. At the learning level, deep networks regress absolute amplitudes without boundary anchoring. On \mbox{sub-metre} gaps the resulting offset makes them less accurate than a straight line between the two measured ends, whereas the straight line ignores lithology when a gap spans several beds, and the best estimator therefore changes with gap length \cite{gama2025imputation}. At the fusion level, classical \mbox{state-space} smoothers weight priors against observations \cite{kalman1960new,rauch1965maximum}, yet Gaussian likelihoods track surviving spikes \cite{zhu2021novel} and stationary covariances \mbox{over-smooth} anchors under strong persistence \cite{durbin2012time}. Moreover, \mbox{cross-log} relations shift across collections \cite{gama2025imputation} whereas local borehole continuity remains informative across basins. Overcoming spike distortion, \mbox{short-gap} boundary offsets and \mbox{cross-basin} degradation simultaneously requires decoupling corruption suppression, measurement anchoring and \mbox{cross-property} inference within one estimator without \mbox{per-scale} switching or retraining.

Overcoming these failures directly benefits quantitative subsurface workflows. Boundary offsets or unstripped spikes in reconstructed density and sonic logs distort acoustic impedance, shift well--seismic ties \cite{zhang2020seismic} and bias porosity, \mbox{pore-pressure} and geomechanical models \cite{chen2020physics,mehrabi2026dual,tassone2019geological}. Maintaining separate \mbox{gap-length} rules or retraining networks per legacy archive is likewise impractical across frontier basins. A \mbox{corruption-resilient}, \mbox{scale-seamless} and \mbox{zero-shot} transferable estimator unifies routine gap completion, \mbox{legacy-log} quality control and sparsely logged well evaluation.

To establish this paradigm, this study develops \mbox{boundary-gated} \mbox{soft-thresholding} Kalman fusion (\mbox{ST-KF}, Fig. 1) around three contributions matching the three bottlenecks. First, multiresolution wavelet soft shrinkage with learnable thresholds suppresses predictor noise while \mbox{inference-time} firm despiking strips isolated spikes without blunting thin beds. Second, a \mbox{distance-gated} residual bridge reformulates Transformer prediction into a continuous correction of the observed chord under training--inference decoupling, training on uncleaned spiky chords to prevent shortcut collapse near boundaries. Third, \mbox{variance-decoupled} robust Kalman smoothing solves conditional residual fusion as an exact \mbox{closed-form} tridiagonal system with firm innovation shrinkage, anchoring outputs to clean measurements while relaxing to \mbox{cross-log} priors across long gaps. Evaluation across public North Sea and Taranaki Basin benchmarks demonstrates \mbox{rank-first} accuracy, flat hyperparameter stability and \mbox{zero-shot} transferability, revealing the decoupling between \mbox{cross-basin} borehole continuity and \mbox{domain-sensitive} \mbox{cross-property} mappings.

\begin{figure*}[!t]

\markrevisedfloat

\centering

\includegraphics[width=\textwidth,height=0.77\textheight,keepaspectratio]{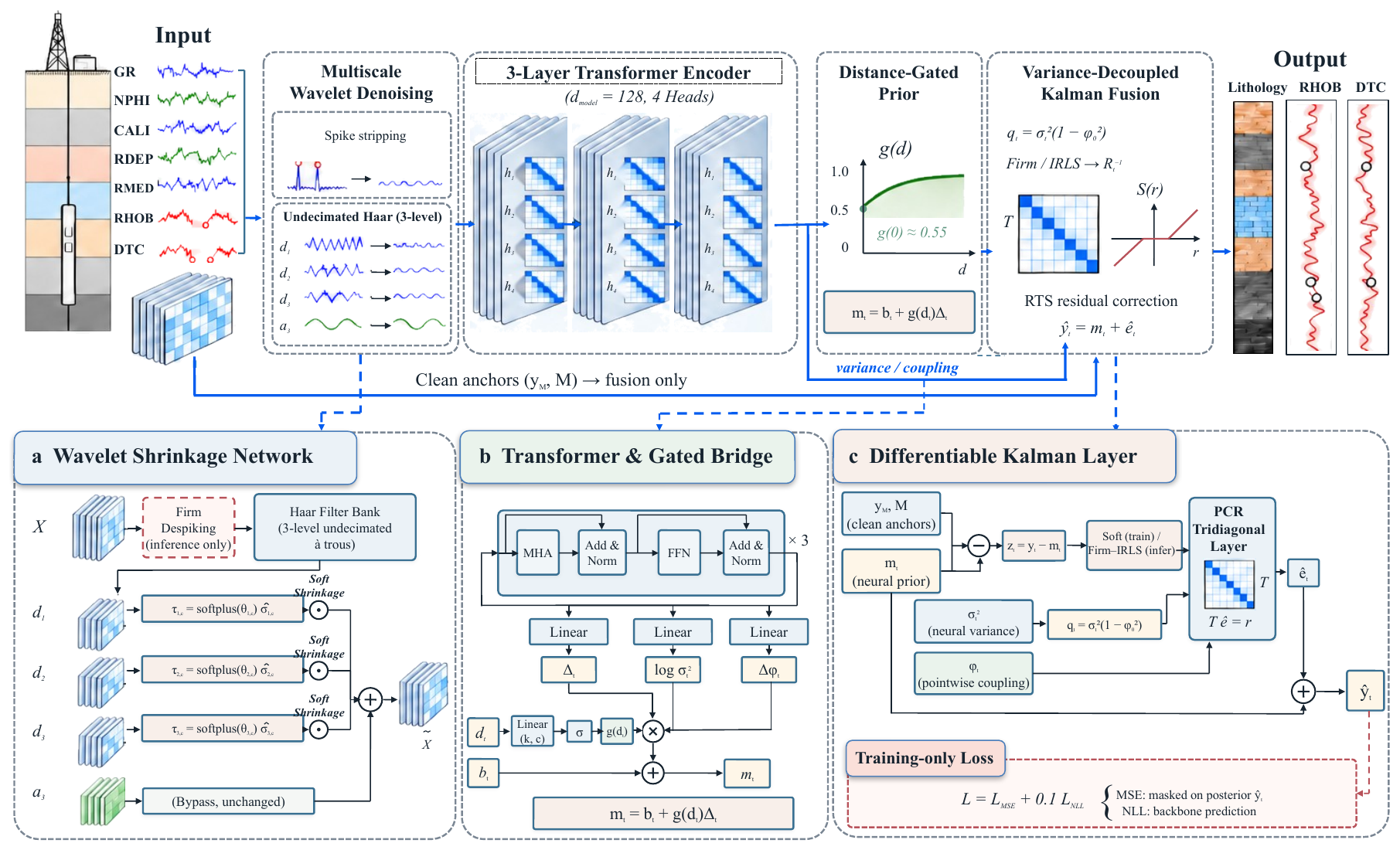}

\caption{\rev{Architecture of \mbox{ST-KF} and its training--inference decoupling. (a) Multiscale wavelet denoising of predictor logs and measured anchors. (b) \mbox{Distance-gated} anchor bridge on the Transformer backbone. (c) \mbox{Soft-thresholding} Kalman fusion solved in closed form. Red dashed boxes mark inference-only despiking (a) and training-only loss supervision (c).}}

\label{fig:1}

\end{figure*}

\section{Related Work and Theoretical Background}

\label{h-s2}

\subsection{\mbox{Well-Log} Reconstruction Methods}

\label{h-s2a}

\mbox{Well-log} reconstruction has followed three routes, and none of them treats the measured samples next to a gap as constraints on the output. The first route regresses the target from \mbox{co-located} logs. Gardner et al.~(1974) provided the earliest substitute with an empirical density--velocity relation \cite{gardner1974formation}, which assumes a fixed lithology and fails across formation boundaries. Chen et al.~(2006) synthesized logs with ensembles of neural networks \cite{chen2006ensemble}, Feng et al.~(2021) imputed missing values with random forests \cite{breiman2001random} and quantified their uncertainty \cite{feng2021rf}, and Kim and Cho (2024) cast imputation as \mbox{nearest-neighbour} collaborative filtering \cite{kim2024knn}. Garini et al.~(2025) showed that gradient boosting fills up to 30\% missing density and sonic values well enough for subsequent lithology classification \cite{garini2025filling}, and Joshi et al.~(2021) compared several learners for sonic prediction \cite{joshi2021prediction}. Zhao et al.~(2025) extended \mbox{machine-learning} regression to seismic anisotropy estimation from vertical seismic profile attributes in an offshore carbonate field \cite{zhao2025anisotropy}. These \mbox{point-wise} models ignore depth context, so they cannot follow beds that the predictor logs resolve only over several samples.

The second route develops temporal modeling for \mbox{well-log} reconstruction, with depth serving as the sequence coordinate. Pham and Zabihi Naeini (2019) cascaded a bidirectional long \mbox{short-term} memory (LSTM) network \cite{hochreiter1997lstm} with dense layers \cite{pham2019missing}, and Zhang et al.~(2018) generated complete synthetic logs with recurrent networks \cite{zhang2018lstm}. Zeng et al.~(2020) added attention to a bidirectional gated recurrent unit (GRU) \cite{cho2014gru} network \cite{zeng2020attention}, Zhou et al.~(2025) stacked bidirectional networks in cascade \cite{zhou2025missing}, and Wang et al.~(2022) combined \mbox{self-attention} networks in an ensemble \cite{wang2022missing}. Li et al.~(2022) predicted compressional and shear slowness jointly with deep sequence learning \cite{li2022well}, Kim (2022) generated shear slowness with hybrid \mbox{machine-learning} models \cite{kim2022synthetic}, and Zhang et al.~(2024) filtered training samples by a formation lithology index before reconstructing distorted logs with an LSTM in this journal's letters \cite{zhang2024well}. Transformers followed for \mbox{drilling-log} quality control in Kumar (2021) \cite{kumar2021transformer} and for sonic prediction in Rahman et al.~(2025) \cite{rahman2025sonic}, while Jiang et al.~(2022) modelled the logs as a gated graph \cite{jiang2022gnn} and Wang et al.~(2026) tuned a hybrid network by Bayesian optimization \cite{wang2026sonic}. Deep models on well logs have become common in this journal more broadly, for example for lithology interpretation \cite{liu2022quantum} and fracture identification \cite{dong2024fisy}. Recurrent restoration has also been applied to other downhole records such as \mbox{gas-well} pressures \cite{ki2020restoration}.

The third route conditions the network on the observed part of the target. Lin et al.~(2023) reconstructed logs missing because of borehole conditions with a sequence \mbox{self-attention} framework \cite{lin2023missing}, Meng et al.~(2024) sampled missing logs from a conditional diffusion model \cite{meng2024missing}, and Guo et al.~(2026) constrained a generative adversarial network with geological information \cite{guo2026gan}. Ali et al.~(2026) added uncertainty estimates through \mbox{physics-guided} ensembles \cite{ali2026physics}. Gama et al.~(2025) assembled a benchmark covering the Geolink, Taranaki and Teapot Dome collections and found that imputation networks outperform interpolation on long gaps but not uniformly \cite{gama2025imputation}. Even on this route, the observed target enters only as an input channel, so nothing forces the output to agree with the anchors at the gap boundary. Most studies on all three routes also evaluate a few wells and one gap regime and report a single training run.

\subsection{Theoretical Foundations of \mbox{Domain-Informed} \mbox{State-Space} Fusion}

\label{h-s2b}

Overcoming boundary drift, spike sensitivity and domain fragility requires unifying multivariate sequence imputation, exact \mbox{state-space} smoothing, wavelet shrinkage and \mbox{domain-informed} regularization within one estimator. General sequence imputation methods treat missing samples as latent variables organized by architecture and missingness pattern \cite{wang2024deep,khayati2024imputevis}, including bidirectional recurrent dynamics in BRITS \cite{cao2018brits}, diagonally masked \mbox{self-attention} in SAITS \cite{du2023saits} and conditional \mbox{score-based} diffusion \cite{tashiro2021csdi}. Efficient Transformer variants reduce attention cost \cite{zhou2021informer,kitaev2020reformer} or introduce frequency and \mbox{trend-seasonal} decomposition \cite{zhou2022fedformer,wu2021autoformer} over the standard Transformer \cite{vaswani2017attention}. Yet these models predict through unconstrained heads without boundary anchoring, and Autoformer strips \mbox{moving-average} trends inside every encoder layer, discarding the \mbox{low-frequency} compaction and lithology baselines essential to wireline logs (Section \mbox{V-A}).

\mbox{State-space} smoothing offers the missing ingredient, namely an exact rule for weighting observations against a prior. For a linear Gaussian \mbox{state-space} model the Kalman filter \cite{kalman1960new} and the RTS smoother \cite{rauch1965maximum} give the exact posterior of the latent state, and for a scalar AR(1) state the smoother reduces to a symmetric tridiagonal linear system \cite{durbin2012time}. Such systems can be solved in \(O(L\log L)\) work and \(O(\log L)\) parallel depth by cyclic reduction \cite{hockney1965fast,zhang2010fast}, which keeps the smoother differentiable and fast on graphics processors. Gaussian smoothing is, however, sensitive to outliers. Huber's \mbox{M-estimation} \cite{huber1964robust}, Hampel's \mbox{bounded-influence} identifiers \cite{hampel1974influence} and the robust Bayesian filter of Masreliez and Martin (1977) \cite{masreliez1977robust} replace or clip outlying innovations before the update. Later work models the noise as \mbox{heavy-tailed}. Jia et al.~(2020) used a Gaussian--Student's \(t\) mixture estimated by variational Bayes \cite{jia2020novel}, Zhu et al.~(2021) handled unknown \mbox{non-stationary} \mbox{heavy-tailed} noise \cite{zhu2021novel}, and Liang et al.~(2025) and Wang et al.~(2026) extended this to \mbox{Pearson-type} mixtures and to \mbox{heavy-tailed} process noise \cite{liang2025robust,wang2026robust}. Wu et al.~(2024) applied such a filter to navigation data fusion \cite{wu2024gaussian}. Learned variants such as deep Kalman filters \cite{krishnan2015deep} and KalmanNet \cite{revach2022kalmannet} estimate transitions or gains with networks. All of these filters assume a known or learned dynamic model of the state, whereas a log gap needs a prior that changes with lithology and is supplied by the remaining logs. \mbox{ST-KF} therefore keeps the exact smoother and lets the network supply the prior mean, the prior variance and the transition coefficient.

Multiresolution wavelet analysis \cite{mallat1989theory} underlies shrinkage denoising, and soft thresholding of wavelet coefficients is \mbox{near-minimax} for sparse signals in Gaussian noise \cite{donoho1994ideal,donoho1995denoising,donoho1995adapting}. Firm thresholding interpolates between hard and soft rules and leaves large coefficients unbiased \cite{gao1997waveshrink}, and undecimated (à trous) transforms avoid the shift dependence of decimated wavelets \cite{shensa1992discrete,coifman1995translation}. Honório et al.~(2012) applied discrete wavelet thresholding to well logs and showed that the choice of threshold decides whether thin beds survive denoising \cite{honorio2012well}. Yu et al.~(2019) replaced \mbox{model-based} denoising of seismic data with deep networks \cite{yu2019denoising}, and Singh et al.~(2020) automated log editing and feature selection with machine learning \cite{singh2020automated}. Pang et al.~(2026) showed that missing and abnormal log responses degrade downstream lithology identification unless the network accounts for data quality \cite{pang2026robust}. In these studies denoising is a separate preprocessing stage, so a spike that survives it still propagates into any reconstruction that trusts the anchors. \mbox{ST-KF} uses soft shrinkage with learnable thresholds on the predictor logs and firm shrinkage on target innovations within the same estimator.

\mbox{Physics-informed} neural networks penalize the residual of governing equations \cite{raissi2019physics,karniadakis2021physics}. Broader taxonomies of \mbox{theory-guided} learning distinguish this from models that embed domain structure such as smoothness, conservation or measurement models \cite{karpatne2017theory,willard2022integrating,reichstein2019deep}. In well logging, Chen and Zhang (2020) constrained the prediction of geomechanical logs with physical relations in this journal \cite{chen2020physics}, and Zhao et al.~(2023) guided \mbox{shear-sonic} prediction with a \mbox{rock-physics} model \cite{zhao2023rock}. \mbox{ST-KF} belongs to the second group. Its prior knowledge is a measurement model (anchors are noisy observations of the same log), a \mbox{depth-correlation} model (AR(1) residuals) and a sparsity model for outliers, not a \mbox{wave-propagation} or \mbox{rock-physics} law. The framework is therefore described as a \mbox{domain-informed} robust \mbox{state-space} model, and it can be combined with \mbox{rock-physics} constraints on the backbone.

\section{Study Area, Data and Evaluation Protocol}

\label{h-s3}

\subsection{Study Area}

\label{h-s3a}

The main study area is the northern North Sea on the Norwegian Continental Shelf, covered by the public FORCE 2020 \mbox{machine-learning} competition release \cite{bormann2020force}. Its 118 wells penetrate a succession that spans the Cenozoic Nordland, Hordaland and Rogaland groups, the Upper Cretaceous Shetland and Lower Cretaceous Cromer Knoll groups, Jurassic groups including the Viking, Tyne, Boknfjord, Vestland (Brent), Dunlin and Båt groups, the Triassic Hegre Group and, in a few wells, the Permian Zechstein and Rotliegendes groups \cite{isaksen1989revised,vollset1984revised} (Fig. 2a, c). These siliciclastic, \mbox{carbonate-bearing} and locally evaporitic successions produce large contrasts in density and slowness at group boundaries, so a reconstruction must reproduce sharp interfaces as well as gradual compaction trends.

The second study area, used only to test transfer, is the Taranaki Basin of New Zealand, a Cretaceous--Cenozoic rift and \mbox{passive-margin} to \mbox{convergent-margin} basin whose stratigraphy is unrelated to that of the northern North Sea \cite{king1996cretaceous} (Fig. 2b). The two basins differ in lithology, burial history and logging contractors, so the Taranaki wells test domain shift rather than interpolation between neighbouring wells.

\subsection{Data and Well Partition}

\label{h-s3b}

Both datasets are public, so that well locations, stratigraphy and raw logs can be inspected and every result can be reproduced. The FORCE 2020 logs are distributed under the Norwegian Licence for Open Government Data, sampled every 0.1524 m and annotated with lithostratigraphic group and formation tops. Five predictor logs, gamma ray (GR), neutron porosity (NPHI), caliper (CALI) and the logarithms of deep and medium resistivity (RDEP and RMED), are used to reconstruct RHOB and DTC. No laboratory core samples were used, and all quantities are wireline measurements.

Cleaning was fixed before any model training. Depth breaks larger than 0.5 m split wells into segments. Physically impossible values, stuck caliper arms and \mbox{flat-lined} curves were set to missing. Isolated one- or \mbox{two-sample} density spikes were removed with a Hampel identifier, while thicker beds such as coals were kept. Density samples with \(|\mathrm{DRHO}|>0.15\) g/cm\(^3\) or caliper more than 2 in above bit size were excluded from loss and evaluation but kept in the input sequence. After cleaning, 114 wells contain at least 1000 jointly valid samples, with 824 905 evaluable RHOB and 859 064 evaluable DTC samples.

Wells sharing a name stem or lying within 3 km of each other were grouped so that no group crosses splits. The partition has 66 training wells, 10 validation wells for early stopping and all design choices, and three test sets (Fig. 2a). The \mbox{in-distribution} blind set has 18 wells drawn by stratified sampling on \mbox{predictor-log} statistics and stratigraphy. The spatially blind set has 10 wells selected by a fixed rule that requires 8--35 km distance to every \mbox{non-blind} well and input statistics within the training support. The official hidden set consists of the 10 hidden wells of the competition, adopted unchanged as a set the authors could not choose. The configuration of \mbox{ST-KF} was frozen on the validation wells before the test sets were read, and each test set was evaluated once.

For the Taranaki Basin, the preprocessed collection of the \mbox{well-log} imputation benchmark \cite{gama2025imputation} is used without any retraining. Of its 407 wells, 149 pass the same unit and coverage checks as the FORCE wells and contain the five predictors and at least one target.

\subsection{Evaluation Protocol}

\label{h-s3c}

Gaps are imposed only on observed target samples, with fixed random seeds shared by all methods (Fig. 2d). The standard set has 14 scenarios that span the gap types met in field logs. Point gaps of one to three samples (0.15--0.46 m), short gaps of 1, 3, 5 and 10 m and long gaps of 20, 50 and 100 m test the dependence on gap length. A natural pattern resamples gap lengths from the empirical distribution of real gaps in the training wells, with about 20\% of samples missing. \mbox{One-sided} gaps of 10 and 50 m have an anchor on only one side, as at the base of a logged interval, and the \mbox{whole-well} scenario removes the target from an entire well. Two spiky versions of the 5- and 20-m scenarios add one- to \mbox{three-sample} spikes of \(\pm3\)--\(6\) standardized units to 2\% of the observed anchors. Eight \mbox{input-noise} scenarios add Gaussian noise with standard deviation 0.1, 0.2, 0.4 or 0.8 standardized units plus 1\% sparse spikes to the five predictor logs, combined with 20-m gaps or \mbox{whole-well} loss. Metrics are computed only on masked, evaluable samples.

Nine baselines cover the main reconstruction routes. All neural baselines were implemented or wrapped by the authors and trained from scratch under identical conditions, with the same cleaned data, splits, 14-channel input (Section \mbox{IV-A}), window length, masks, spike augmentation, early stopping and five seeds. BiGRU, BiLSTM, a \mbox{one-dimensional} \mbox{U-Net} \cite{ronneberger2015unet}, the Transformer encoder (the \mbox{ST-KF} backbone without gate and fusion) and Autoformer \cite{wu2021autoformer} are our own implementations, and SAITS uses the backbone of the PyPOTS library \cite{du2023saits,du2023pypots}. LightGBM \cite{ke2017lightgbm} regresses each target from the five predictors at the same depth, and linear interpolation uses the chord between anchors. The \mbox{initial-submission} model of this study (\mbox{ST-KF-Autoformer}) was retrained in the same setting for reference. Learning rates, software and hardware are listed in Appendix A.

For each well, RMSE and MAE are computed in physical units and averaged over five independently trained seeds. Tables report the mean over wells, and pooled \(R^2\) is computed over all masked samples of a test set. Differences between \mbox{ST-KF} and each baseline are tested per scenario and log with the \mbox{two-sided} Wilcoxon \mbox{signed-rank} test and the paired \(t\)-test on \mbox{well-level} errors \cite{wilcoxon1945individual,demsar2006statistical}, with Holm correction across the baselines \cite{holm1979simple}. \mbox{Ninety-five} percent confidence intervals come from 2000 \mbox{well-level} bootstrap resamples \cite{efron1979bootstrap}.

\begin{figure*}[!t]

\markrevisedfloat

\centering

\includegraphics[width=\textwidth,height=0.77\textheight,keepaspectratio]{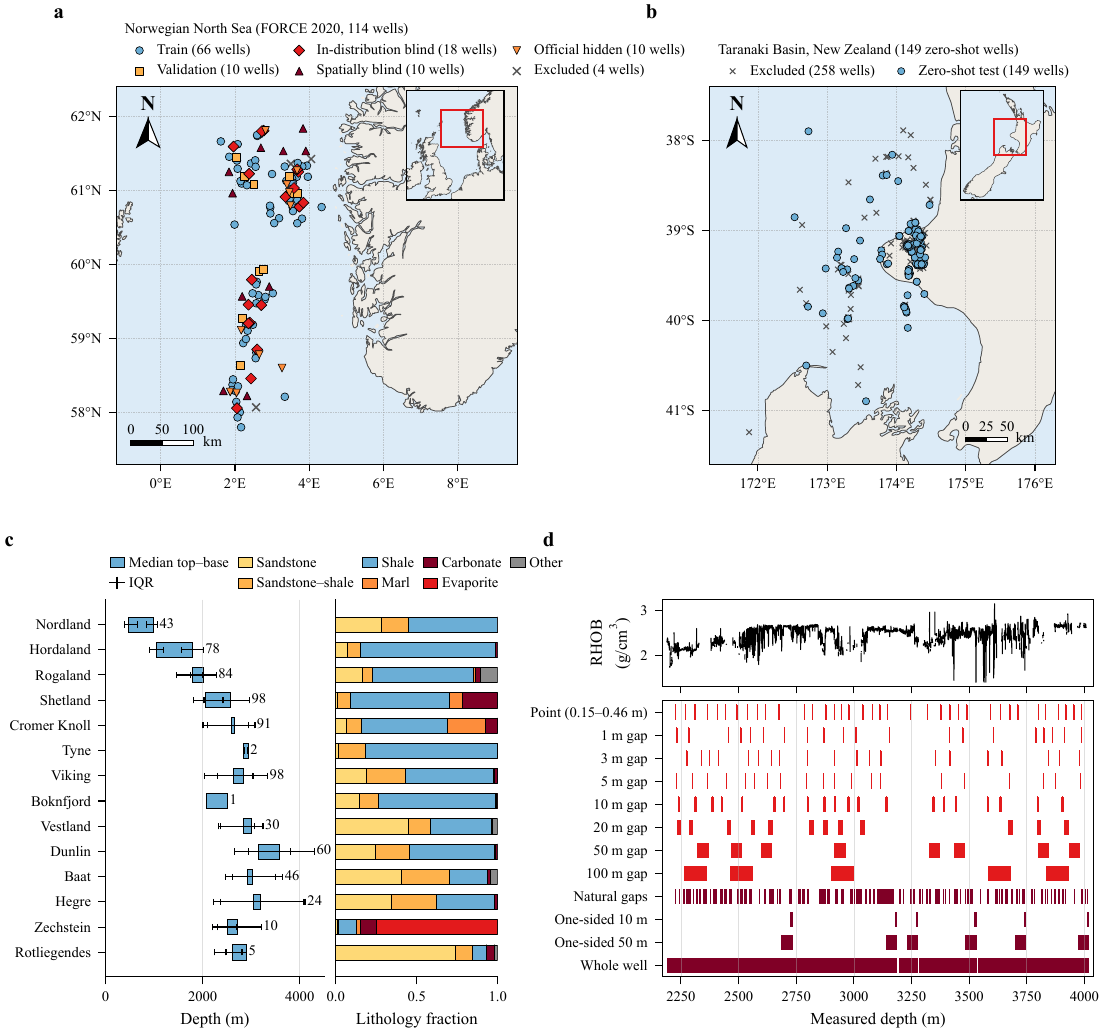}

\caption{\rev{Study areas, well partition and gap protocol used to test whether one model generalizes across wells, regions and gap lengths. (a) FORCE 2020 wells in the northern North Sea coloured by split. (b) Taranaki Basin wells used for \mbox{zero-shot} transfer. Maps use WGS84 coordinates (FORCE locations converted from ED50/UTM zone 31N) and Natural Earth coastlines, and insets show the regional setting. (c) Depth ranges of the lithostratigraphic groups in the FORCE wells. (d) Gap intervals imposed by each scenario on one validation segment.}}

\label{fig:2}

\end{figure*}

\section{The \mbox{ST-KF} Framework}

\label{h-s4}

\subsection{Problem Formulation}

\label{h-s4a}

Let \(\mathbf{x}_t\in\mathbb{R}^5\) be the predictor logs at depth index \(t\), \(y_t\in\mathbb{R}^2\) the standardized target logs and \(M_t\in\{0,1\}^2\) the indicator that \(y_t\) is observed. RHOB and DTC are the two targets in this study, but nothing in the formulation is specific to them, and any log that is measured on part of a well can take their place. Consider a well interval with \(L=1024\) equally spaced samples (156 m), where the reconstruction objective is formulated as estimating \(y_t\) wherever \(M_t=0\). Three \mbox{domain-specific} characteristics of field data shape the estimator. The predictor logs carry tool noise and sparse spikes, and the observed target samples next to a gap, called anchors below, can themselves be spikes. The reconstruction has to meet clean anchors without a step at the gap boundary. The amount of anchor information falls from two close neighbours in a point gap to none in an unlogged well, so the weight given to anchors and to \mbox{cross-log} inference has to change with the distance to the nearest anchor.

The model input has 14 channels (Fig. 1), namely the five predictors, an \mbox{input-validity} flag, the observed targets \(y_tM_t\), the masks \(M_t\), the \mbox{linear-interpolation} chord \(b_t\) built only from observed samples (constant extrapolation beyond the last anchor and zero where no anchor exists), and the distance to the nearest anchor \(d_t=\log(1+\delta_t)/\log 101\), where \(\delta_t\) is the distance in samples. Sliding windows are stitched at inference, and \(b_t\) and \(d_t\) are computed on the whole segment before windowing so that the chord keeps its \mbox{two-sided} slope across window boundaries. All baselines receive the same 14 channels.

\subsection{Overall Architecture}

\label{h-s4b}

Rather than scaling network capacity, \mbox{ST-KF} aligns operator structure with borehole logging mechanics through three sequential modules (Fig. 1). Multiresolution \mbox{soft-thresholding} denoising separates \mbox{high-frequency} noise and isolated spikes from \mbox{low-frequency} stratigraphic trends (Section \mbox{IV-C}). A \mbox{distance-gated} bridge reformulates Transformer prediction into a continuous residual correction of the anchor chord, handing weight smoothly from anchors to \mbox{cross-log} inference with distance (Section \mbox{IV-D}). \mbox{Variance-decoupled} Kalman fusion reconciles this prior with clean anchors in closed form while assigning zero weight to fully stripped spikes (Section \mbox{IV-E}), yielding \(\hat y_t=m_t+\hat e_t\).

Explicit training--inference decoupling shares one frozen weight set across regimes while deploying outlier filters asymmetrically to prevent shortcut learning. Training retains spiky anchors and chords under differentiable soft innovation shrinkage, forcing \mbox{cross-property} learning right up to boundaries. Inference activates input despiking, anchor cleaning and firm shrinkage with reweighting without altering weights, suppressing spikes while leaving clean intervals almost unaffected (Sections \mbox{IV-D} and \mbox{V-B}).

\subsection{Multiscale \mbox{Soft-Thresholding} Denoising}

\label{h-s4c}

Separating sensor noise and spikes from geological signal requires \mbox{frequency-selective} shrinkage, because compaction trends and lithology baselines occupy low frequencies whereas tool noise and thin beds reside in distinct detail bands. A \mbox{three-level} undecimated Haar transform preserves coarse approximations \(a_j(t)=\tfrac12[a_{j-1}(t)+a_{j-1}(t+2^{j-1})]\) while shrinking only detail bands \(d_j=a_{j-1}-a_j\) (\(a_0=\mathbf{x}\)),

\begin{equation}\begin{aligned}\hat d_{j,c}&=\operatorname{sign}(d_{j,c})\max(|d_{j,c}|-\tau_{j,c},0),\\ \tau_{j,c}&=\operatorname{softplus}(\theta_{j,c})\,\hat\sigma_{j,c},\end{aligned}\tag{1}\end{equation}

where \(\hat\sigma_{j,c}=1.4826\,\operatorname{med}|d_{j,c}|\) is Donoho's robust noise scale \cite{donoho1994ideal} and \(\theta_{j,c}\) is learned. The reconstruction is \(a_3+\sum_j\hat d_j\). At inference, isolated input spikes are removed before this transform. With a local median \(b^x\) over 15 samples and a local scale \(s^x\) (\(1.4826\times\) the median absolute deviation over 49 samples), the \mbox{high-pass} residual \(h^x=x-b^x\) is replaced by \(\operatorname{Firm}(h^x;\max(3.8s^x,1.8),\max(6s^x,3))\) only on runs of at most three samples, where

\begin{equation}\operatorname{Firm}(h;\tau_1,\tau_2)=\begin{cases}0,&|h|\le\tau_1\\ \operatorname{sign}(h)\,\tau_2\frac{|h|-\tau_1}{\tau_2-\tau_1},&\tau_1<|h|<\tau_2\\ h,&|h|\ge\tau_2\end{cases}\tag{2}\end{equation}

is the \mbox{firm-threshold} operator \cite{gao1997waveshrink} and the subtracted quantity is the stripped spike. The absolute floors (in standardized units) protect thin beds in quiet shale intervals, where the local scale is small.

The observed anchors are cleaned in the same way at inference. Spikes that the fusion head would strip completely (Section \mbox{IV-E}) are removed from \(y_t\), and only the resulting change of the chord inside the current window is added to \(b_t\). A clean window therefore keeps the global, \mbox{two-sided} chord exactly, and a corrected anchor changes the chord only where the spike was.

\subsection{\mbox{Distance-Gated} Anchor Bridge}

\label{h-s4d}

Eliminating boundary steps requires reformulating the learning target rather than \mbox{post-hoc} blending, because unconstrained amplitude regression leaves systematic offsets at gap edges. On a \mbox{three-layer} Transformer encoder (\(d_{\text{model}}=128\), four heads, \mbox{feed-forward} width 512, sinusoidal positional encoding, 597 793 total parameters), the network predicts a residual \(\Delta_t\), \mbox{log-variance} \(\log\sigma_t^2\) and coupling offset \(\Delta\phi_t\) per target, forming the gated chord prior

\begin{equation}m_t=b_t+g(d_t)\,\Delta_t,\quad g(d)=\sigma(k(d-c)),\tag{3}\end{equation}

with \(k\) and \(c\) learned per target. Near an anchor (\(d_t\to0\)) the prior is pulled toward the chord and the boundary step disappears. Deep inside a long gap \(g\to1\) and \(m_t\) is the network's \mbox{cross-log} inference. \mbox{Whole-well} gaps have \(b_t=0\) and \(m_t=\Delta_t\) up to the gate value. The learned centres are close to zero (between \(-0.05\) and \(0.03\) over five seeds), so \(g\approx0.55\) at the anchor itself, \(g\approx0.8\) one sample (0.15 m) away and \(g\approx0.95\) four samples away. When the chords are cleaned during training, \(g\) one sample away drops to about 0.2 (Fig. 7d). The gate does not interpolate by itself, and exact agreement with clean anchors comes from the fusion step.

Leaving chords \(b_t\) uncleaned during training prevents boundary shortcut collapse. Spiky training anchors corrupt \(b_t\), penalizing naive chord copying and forcing the backbone to extract \mbox{cross-log} features even one sample from an anchor (\(c\approx0\)). \mbox{Pre-cleaning} training chords creates a copying shortcut \cite{geirhos2020shortcut} that traps \(c\) near 0.35 and suppresses \mbox{short-gap} corrections (Section \mbox{V-B}), whereas \mbox{inference-only} cleaning (Section \mbox{IV-C}) strips field spikes without altering the learned gate.

\subsection{\mbox{Soft-Thresholding} Kalman Fusion}

\label{h-s4e}

Tight boundary consistency without tracking corrupted anchors requires \mbox{closed-form} reconciliation between the gated prior and noisy measurements, overcoming the outlier pull and \mbox{high-persistence} stiffness of stationary Gaussian smoothers. \mbox{Soft-thresholding} Kalman fusion models the residual \(e_t=y_t-m_t\) as a \mbox{non-homogeneous} AR(1) process observed at anchors,

\begin{equation}\begin{aligned}e_t&=\phi_t\,e_{t-1}+w_t,& w_t&\sim\mathcal N(0,q_t),\\ z_t&=e_t+o_t+v_t,& v_t&\sim\mathcal N(0,R_t)\ (M_t=1),\end{aligned}\tag{4, 5}\end{equation}

where \(z_t=y_t-m_t\) is the innovation and \(o_t\) a sparse outlier. The coefficient \(\phi_t=\sigma(\phi_{\text{raw}}+\Delta\phi_t)\le0.995\) is predicted \mbox{point-wise} so that coupling can drop at lithological interfaces. The usual stationary choice \(q_t=\sigma_t^2(1-\phi_t^2)\) ties amplitude to persistence. When \(\phi_t\to1\) the process noise vanishes, the smoother becomes stiff and anchors are \mbox{over-smoothed}. \mbox{ST-KF} decouples them,

\begin{equation}q_t=\sigma_t^2\,(1-\phi_0^2),\quad \phi_0\in[0.85,0.985],\tag{6}\end{equation}

with a learned global \(\phi_0\) (0.94--0.96 after training) and \(R\in[10^{-4},10^{-3}]\) in standardized units. The ratio \(R/q_t\) at anchors then stays near one, so the posterior passes through clean anchors and relaxes to the prior within a few samples.

Outliers are estimated once on the \mbox{pre-smoothing} innovation, so that the slack of the smoother next to a gap cannot be mistaken for an outlier. With a 31-sample local median \(\operatorname{med}_t(z)\) and a 65-sample robust scale \(s_t\) computed on observed samples, \(h_t=z_t-\operatorname{med}_t(z)\) and

\begin{equation}\begin{aligned}o_t&=\operatorname{Firm}(h_t;\tau_t,1.8\tau_t)\,\mathbf{1}[\text{run}\le3],\\ \tau_t&=\max(\kappa s_t,\tau_{\min}),\end{aligned}\tag{7}\end{equation}

where \(\kappa\ge1\) is learned, \(\tau_{\min}=1.5\) and the indicator keeps only isolated runs of at most three samples, which separates spikes from beds. One IRLS step then \mbox{down-weights} the corresponding observations, \(R_t^{-1}=R^{-1}(1-|o_t|/|h_t|)^2\), so a fully stripped spike has zero weight. Given \(o_t\), the model is linear and Gaussian in \(e\), and the RTS smoother on the \mbox{pseudo-observations} \(\tilde z_t=z_t-o_t\) is the minimizer of

\begin{equation}\begin{aligned}&\sum_{t:M_t=1}\frac{(\tilde z_t-e_t)^2}{R_t}\\ &\quad+\sum_t\frac{(e_t-\phi_te_{t-1})^2}{q_t}+\frac{e_1^2}{\sigma_1^2},\end{aligned}\tag{8}\end{equation}

whose symmetric tridiagonal normal equations are solved exactly and differentiably in \(\lceil\log_2L\rceil\) parallel steps via \mbox{double-precision} cyclic reduction. The posterior \(\hat y_t=m_t+\hat e_t\) unifies all gap scales analytically from (8). Short gaps propagate clean anchor residuals inward via AR(1) correlation while the backbone supplies bed shape, long gaps decay \(\hat e_t\to0\) toward \(m_t\), and \mbox{whole-well} loss reduces identically to the prior. Differentiable soft shrinkage \(o_t=\operatorname{sign}(h_t)\max(|h_t|-\tau_t,0)\) trains end to end, while firm shrinkage with IRLS suppresses spikes selectively at inference (Section \mbox{IV-B}).

\section{Results}

\label{h-s5}

\subsection{Comparative Experiments}

\label{h-s5a}

\begin{table*}[!t]

\markrevisedfloat

\centering\footnotesize\setlength{\tabcolsep}{2pt}

\caption{\rev{Pooled reconstruction accuracy over the 13 gap scenarios with anchors (\mbox{whole-well} loss excluded) on the three blind test sets. RMSE in g/cm\(^3\) (RHOB) and \(\mu\)s/ft (DTC), mean of five seeds, with best values in bold.}}

\label{tbl-tab1}

\input{tables/r1_tbl-tab1.tex}

\par\smallskip{\footnotesize Note. ‡Model of the initial submission, retrained under the present protocol.\par}

\end{table*}

\mbox{Measurement-guided} boundary coupling and \mbox{closed-form} smoothing give \mbox{ST-KF} \mbox{rank-first} pooled accuracy and \(R^2\) for both targets across all three blind cohorts without enlarging the backbone (Table I). Against the strongest of nine baselines per cohort, pooled RMSE drops by 5.8\%, 5.0\% and 8.3\% for RHOB and by 6.4\%, 12.7\% and 15.6\% for DTC across the \mbox{in-distribution}, spatially blind and official hidden sets. \mbox{ST-KF} ranks first in 150 of 168 evaluation cells (averaging 9.4\% over the \mbox{runner-up}) and wins 216, 204 and 206 of 250 \mbox{Holm-corrected} paired tests per cohort with zero significant losses. Because Transformer and SAITS receive identical anchor inputs yet trail consistently, the gain stems from \mbox{state-space} boundary reconciliation rather than anchor access. \mbox{Short-gap} waveforms illustrate this mechanism (Fig. 3). Linear interpolation cuts across \mbox{thin-bed} extrema and unconstrained networks leave boundary offsets, whereas \mbox{ST-KF} matches endpoints closely while tracking \mbox{predictor-driven} beds, lowering gap RMSE 1.6--4.4-fold in \mbox{high-contrast} cases (0.089 vs 0.214 g/cm\(^3\) vs interpolation in Fig. 3a, 0.034 vs 0.055 g/cm\(^3\) vs Transformer in b, and 0.199 vs 0.873 and 0.221 vs 0.961 \(\mu\)s/ft vs BiGRU in c,d), alongside \mbox{full-cohort} gains in Fig. 4 and Table II.

\begin{figure*}[!t]

\markrevisedfloat

\centering

\includegraphics[width=\textwidth,height=0.77\textheight,keepaspectratio]{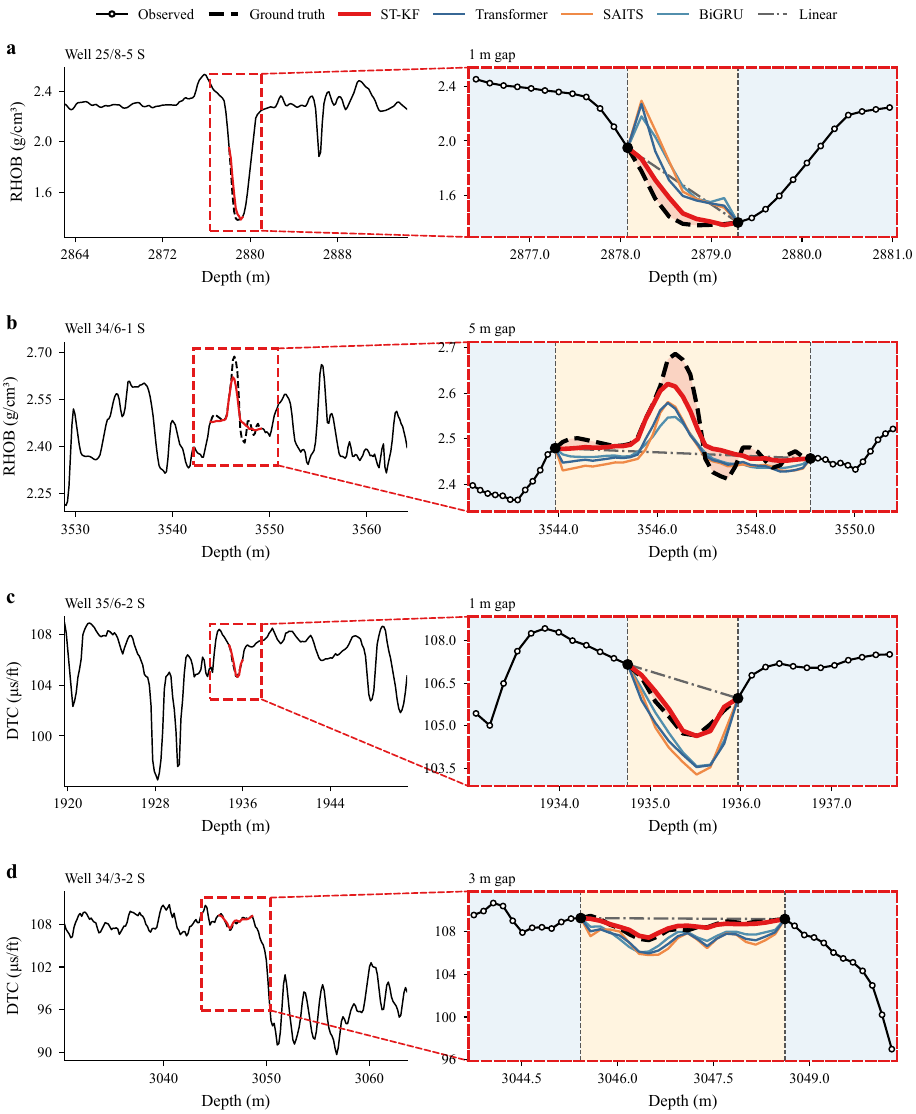}

\caption{\rev{Showcase of how \mbox{ST-KF} reconstructs short gaps between measured anchors in blind wells. Each row shows a 30-m context track with the gap marked by a red dashed box (left) and the magnified gap with its observed neighbourhood (right). Black dots mark the measured anchors and shading separates the observed sections (blue) from the gap (cream). (a) 1-m and (b) 5-m RHOB gaps, (c) 1-m and (d) 3-m DTC gaps. The four cases were selected by a fixed rule as the gaps with the largest advantage of \mbox{ST-KF} over all baselines (top 0.5\% of 3638 candidate gaps), and representative statistics are given in Fig. 4 and Table II.}}

\label{fig:3}

\end{figure*}

\begin{figure*}[!t]

\markrevisedfloat

\centering

\includegraphics[width=\textwidth,height=0.77\textheight,keepaspectratio]{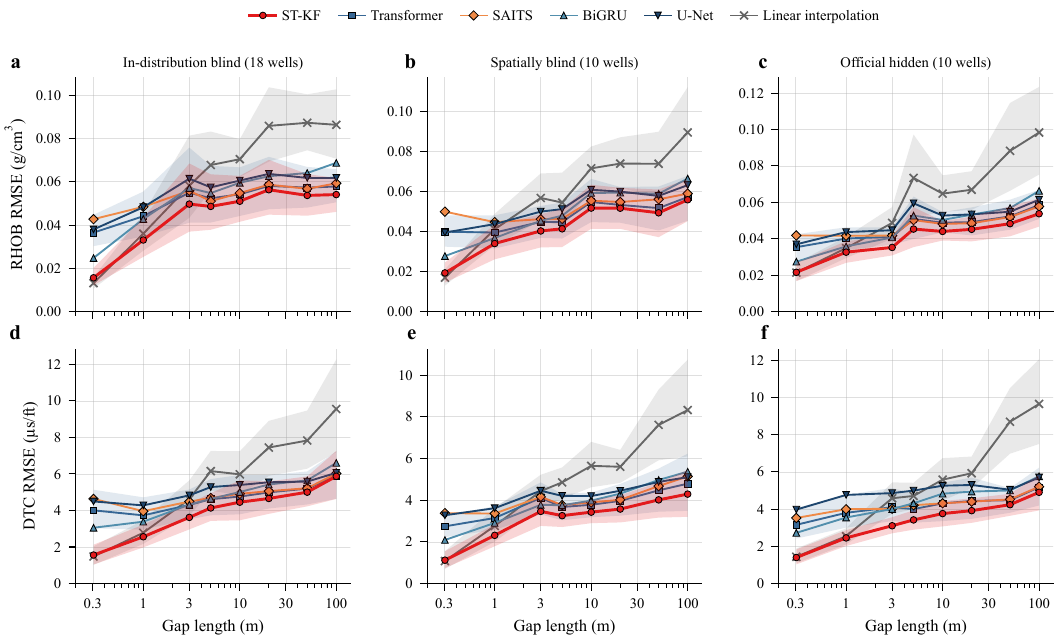}

\caption{\rev{Dependence of reconstruction error on gap length for the three blind test sets. Rows show RHOB and DTC, and shaded bands are 95\% \mbox{well-level} bootstrap intervals. Gap lengths range from one to three samples (0.15--0.46 m) to 100 m.}}

\label{fig:4}

\end{figure*}

Across gap lengths from 0.15 m to 100 m, \mbox{ST-KF} eliminates the classical crossover between \mbox{short-gap} interpolation and \mbox{long-gap} regression (Fig. 4, Table II). Error grows smoothly below every neural baseline across all anchored scales, whereas linear interpolation doubles its error from 3 m to 100 m. Gains peak on \mbox{short-to-medium} intervals, reaching 3-m \mbox{official-hidden} RMSE of 0.0352 g/cm\(^3\) and 3.10 \(\mu\)s/ft versus 0.0407 g/cm\(^3\) and 4.00 \(\mu\)s/ft for the best neural baseline (Table II). Linear interpolation competes only on 0.15--0.46-m point gaps with immediate 0.15-m neighbours, edging \mbox{ST-KF} \mbox{non-significantly} on the \mbox{in-distribution} and spatially blind sets (RHOB 0.0132 vs 0.0156 and 0.0169 vs 0.0193 g/cm\(^3\), DTC 1.48 vs 1.56 and 1.08 vs 1.12 \(\mu\)s/ft, \(p_{\text{Holm}}\ge0.08\)). From 1 m upwards \mbox{ST-KF} achieves the lowest mean error in all six cells, and from 3 m upwards every improvement over interpolation is significant.

\begin{table*}[!t]

\markrevisedfloat

\centering\footnotesize\setlength{\tabcolsep}{2pt}

\caption{\rev{\mbox{Scenario-level} RMSE on the official hidden wells (mean of five seeds). * and † mark \mbox{ST-KF} as significantly better or worse than that baseline (\mbox{well-paired} Wilcoxon test, \mbox{Holm-corrected} \(p<0.05\)), and best values are in bold. Noise level in standardized units of the predictor logs. The other two test sets are in Supplementary Tables S1--S2.}}

\label{tbl-tab2}

\input{tables/r1_tbl-tab2.tex}

\end{table*}

\mbox{Whole-well} and decomposition comparisons confirm the necessity of preserving \mbox{low-frequency} stratigraphic baselines. Without anchors, \mbox{ST-KF} reduces to its Transformer prior, achieving the lowest \mbox{whole-well} RMSE on \mbox{in-distribution} RHOB, DTC and \mbox{official-hidden} DTC while trailing the best of Transformer, BiGRU and SAITS by an insignificant 1.2\%--4.6\% elsewhere. Conversely, \mbox{series-decomposition} models (Autoformer and initial \mbox{ST-KF-Autoformer}) degrade to pooled \mbox{in-distribution} DTC RMSE of 21.4 and 18.2 \(\mu\)s/ft versus 5.37 \(\mu\)s/ft for \mbox{ST-KF}, because encoder trend removal strips away burial compaction and lithology baselines.

\begin{figure*}[!t]

\markrevisedfloat

\centering

\includegraphics[width=\textwidth,height=0.77\textheight,keepaspectratio]{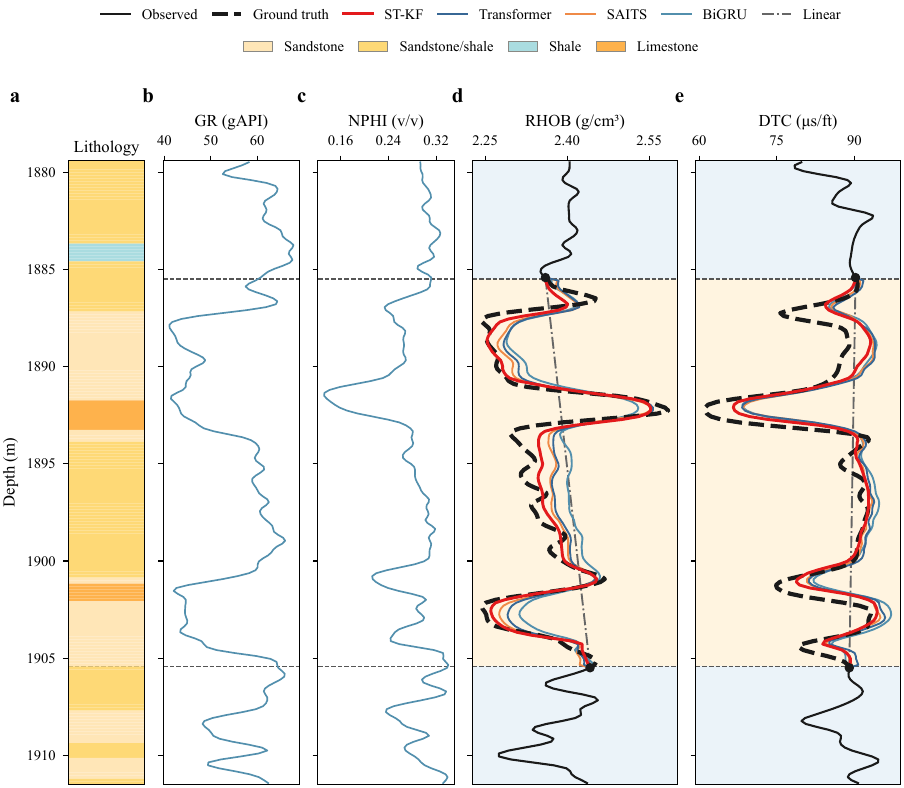}

\caption{\rev{Reconstruction of a 20-m gap through interbedded sandstone, shale and limestone in blind well 31/3-3 (1885.6--1905.3 m). Tracks show (a) lithology, (b) gamma ray, (c) neutron porosity, (d) RHOB and (e) DTC. In (d) and (e) the observed sections are shaded blue, the gap cream, and black dots mark the gap anchors. The gap was selected by a fixed rule among 273 candidate 20-m gaps. Five gaps in which \mbox{ST-KF} has \(R^2\ge0.7\) and the lowest error on both logs qualified, and the one with the most varied lithology was taken.}}

\label{fig:5}

\end{figure*}

Across a 20-m interbedded sandstone, shale and limestone interval (Fig. 5), \mbox{ST-KF} matches both anchors while resolving \mbox{gamma-ray} and \mbox{neutron-porosity} bed alternations, reaching gap RMSE of 0.026 g/cm\(^3\) and 4.11 \(\mu\)s/ft versus 0.035--0.054 g/cm\(^3\) and 4.86--5.13 \(\mu\)s/ft for neural baselines and 0.093 g/cm\(^3\) and 8.95 \(\mu\)s/ft for interpolation. Remaining errors across all models localize at sharp bed boundaries and thin carbonate peaks, matching the 20-m cohort statistics in Table II and Fig. 4.

\begin{figure*}[!t]

\markrevisedfloat

\centering

\includegraphics[width=\textwidth,height=0.77\textheight,keepaspectratio]{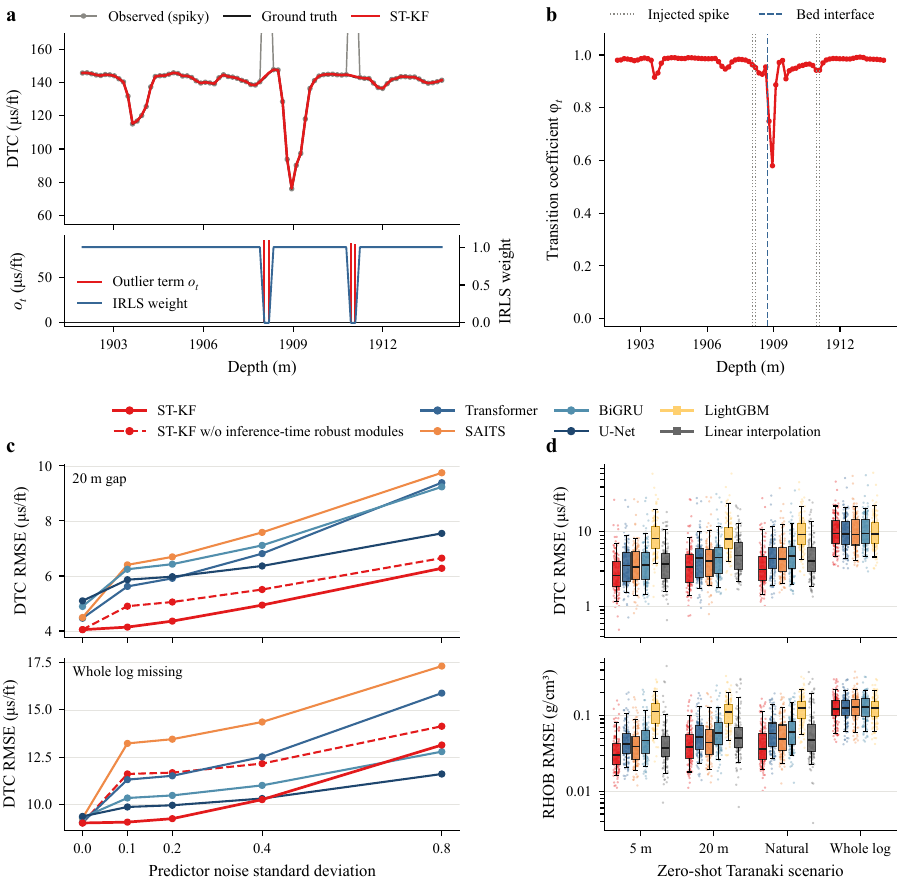}

\caption{\rev{Reconstruction under corrupted anchors, noisy predictor logs and a new basin. (a) Spiky anchors next to a gap and the firm outlier term. (b) \mbox{Point-wise} transition coefficient along the same interval, with injected spikes and the largest true step marked. (c) DTC RMSE versus \mbox{predictor-noise} level for 20-m and \mbox{whole-well} gaps, averaged over the three blind test sets. (d) \mbox{Well-level} RMSE in the Taranaki Basin without \mbox{fine-tuning}.}}

\label{fig:6}

\end{figure*}

Separating sensor spikes from bed boundaries protects fusion against anchor anomalies that more than double interpolation error (Table II). \mbox{ST-KF} strips 94.6\% of validation DTC anchor spike energy versus 30\% without inference modules, ranking first in every spiky test cell with 3.5\%--11.1\% lower DTC error. Adjacent to a spiky gap (Fig. 6a,b), firm thresholding captures nearly full spike amplitudes into \(o_t\) (87--91 of 88--91 \(\mu\)s/ft) and IRLS zeros weights on fully stripped spikes, keeping \(\phi_t\) high (0.94--0.96) across spikes while dropping to 0.6--0.75 at genuine lithological steps.

Multiresolution input shrinkage sustains accuracy under degraded predictors inside long gaps. Across 0.1--0.4 standardized predictor noise, \mbox{ST-KF} ranks first in 34 of 36 cells and cuts error by 8.6\%--34.4\% over disabling the inference \mbox{front-end} (Fig. 6c). Only under extreme 0.8 dense Gaussian noise with unanchored \mbox{whole-well} loss do local filters edge ahead on DTC, where \mbox{U-Net} improves by 8.5\%--16\% (\(p_{\text{Holm}}=0.039\) on official hidden, the sole significant loss among 942 comparisons) and BiGRU averages 12.8 vs 13.1 \(\mu\)s/ft, indicating an advantage for local convolutional or recurrent filtering in this unanchored regime.

\begin{table*}[!t]

\markrevisedfloat

\centering\footnotesize\setlength{\tabcolsep}{2pt}

\caption{\rev{Computational cost on one NVIDIA RTX 5080, including parameters, training time per seed (mean \(\pm\) s.d. over five seeds) and inference latency per well (18 \mbox{in-distribution} wells, 20-m gaps), with the mean pooled DTC RMSE of Table I.}}

\label{tbl-tab5}

\input{tables/r1_tbl-tab5.tex}

\par\smallskip{\footnotesize Note. ¶Measured in a second session together with the same weights without \mbox{inference-time} modules (20.0 ms) and the Transformer (6.0 ms), and other latencies are from the first session. Training times were recorded on a shared GPU and indicate magnitude only.\par}

\end{table*}

These gains incur minimal overhead (Table III). \mbox{ST-KF} adds just 543 parameters (0.09\%) over the Transformer and runs at 22.4 ms total latency per well (3.7 times the \mbox{same-session} Transformer, with inference modules adding 12\% over the 20.0-ms \mbox{module-free} network, with 13.9-min seed training from \mbox{double-precision} solving), matching SAITS and BiLSTM latency at much lower error to process hundreds of wells in seconds.

\subsection{Ablation Experiments}

\label{h-s5b}

\begin{table*}[!t]

\markrevisedfloat

\centering\footnotesize\setlength{\tabcolsep}{2pt}

\caption{\rev{Ablation on the three blind test sets, given as the mean change of RMSE (\%) when a component is removed or replaced (positive values mean worse without the component), averaged over test sets and over the scenarios of each group. (a) \mbox{Inference-time} modules, relative to \mbox{ST-KF}. (b) Trained components, relative to the same network without \mbox{inference-time} modules, with each variant was retrained with five seeds.}}

\label{tbl-tab4}

\input{tables/r1_tbl-tab4.tex}

\par\smallskip{\footnotesize Note. §Built on the earlier fusion head with coupled \(q_t\), and the row includes that head's \mbox{short-gap} deficit.\par}

\end{table*}

Ablations confirm orthogonal operator synergy (Table IV). Relative to the trained network without inference modules (Table IVb), boundary and \mbox{state-space} terms dominate short gaps. Removing the \mbox{distance-gated} bridge increases the error on gaps of 1 m or less by 9.7\% (RHOB) and 15.9\% (DTC) and on 3--100-m gaps by 4.8\% and 8.0\%. Coupled variance \(q_t=\sigma_t^2(1-\phi_t^2)\) with \mbox{smoothed-residual} outlier testing increases \mbox{sub-metre} error by 11.7\% and 31.6\% via covariance stiffness (\(R/q\) in the hundreds) and seed variance. Removing input wavelet shrinkage increases \mbox{noisy-input} error by 15.7\% and 10.2\%, and removing Kalman fusion increases \mbox{sub-metre} error by 38.7\% and 58.8\%. \mbox{Whole-well} impacts stay within \(\pm2\%\), while disabling inference modules relative to full \mbox{ST-KF} (Table IVa) shifts clean error by at most 0.2\% versus 5.7\% \mbox{spiky-DTC} and 26.4\% and 16.5\% \mbox{noisy-input} penalties.

Causal validation confirms that training--inference decoupling prevents shortcut collapse (Fig. 7d). \mbox{Pre-cleaning} training chords shifts the gate centre to \(c=0.28\)--0.39 versus \(-0.05\) to \(0.03\) in \mbox{ST-KF}, raising \mbox{point-gap} DTC RMSE from 1.691 to 1.917 \(\mu\)s/ft. Fixing \(c\) to the \mbox{ST-KF} value restores \mbox{point-gap} error to 1.696 \(\mu\)s/ft and recovers half the overall DTC gap (4.909 vs 5.037 and 4.778), proving a \mbox{gate-mediated} copying shortcut alongside stronger \mbox{cross-log} features learned from noisy chords.

\subsection{Sensitivity Experiments}

\label{h-s5c}

\begin{figure*}[!t]

\markrevisedfloat

\centering

\includegraphics[width=\textwidth,height=0.77\textheight,keepaspectratio]{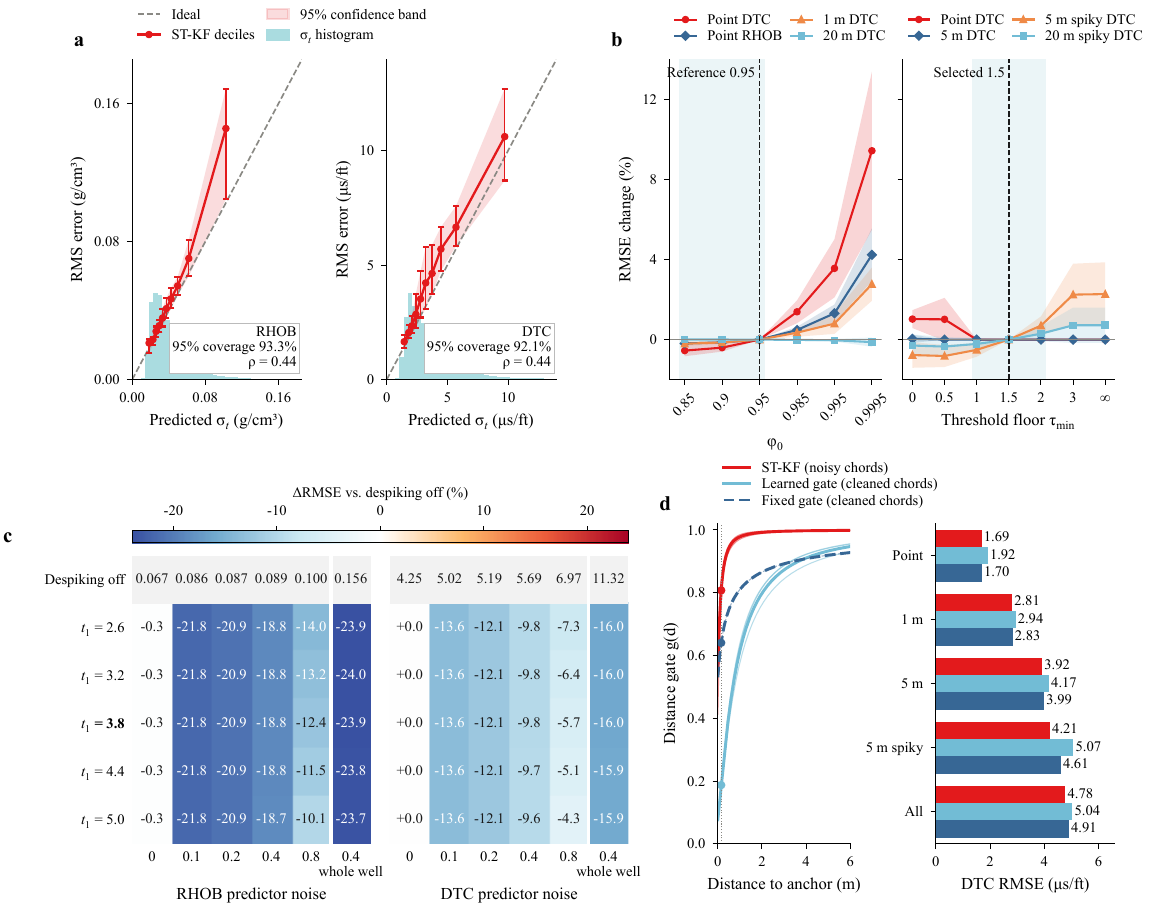}

\caption{\rev{Reliability of the predicted uncertainty and stability of the frozen configuration. (a) Backbone standard deviation versus empirical RMS error on masked samples of the \mbox{in-distribution} blind wells, binned by deciles of \(\sigma_t\), with 95\% \mbox{well-bootstrap} intervals and the distribution of \(\sigma_t\). (b) Sensitivity to the persistence \(\phi_0\) and the threshold floor \(\tau_{\min}\) on the validation wells (mean \(\pm1\) s.d. over seeds, flat region shaded, network without \mbox{inference-time} modules). (c) RMSE change of 20-m and \mbox{whole-well} reconstruction relative to disabled input despiking, for each despiking threshold \(t_1\) and \mbox{predictor-noise} level (validation wells, mean of three seeds), with the absolute RMSE without despiking in the top row and the frozen \(t_1\) in bold. (d) Learned distance gate \(g\) for every seed and the resulting DTC RMSE for noisy and cleaned training bridges and for the \mbox{fixed-gate} control (validation wells), with dots marking \(g\) one sample from the anchor.}}

\label{fig:7}

\end{figure*}

Calibration and sensitivity tests confirm reliable error screening and \mbox{zero-retuning} stability (Fig. 7). Prior uncertainty \(\sigma_t\) tracks empirical error monotonically across deciles (Fig. 7a), while validation RMSE varies under 1\% across \(\phi_0\in[0.85,0.95]\) and \(\tau_{\min}\in[1,2]\) (Fig. 7b). Input despiking cuts 20-m error by 18.8\%--21.8\% (RHOB) and 9.8\%--13.6\% (DTC) at 0.1--0.4 noise and \mbox{whole-well} 0.4-noise error by 23.9\% and 16.0\% within 0.3\% clean shift, with frozen \(t_1=3.8\) preserving thin beds over aggressive 0.8-noise smoothing (\(-5.7\%\) vs \(-7.3\%\) at \(t_1=2.6\), Fig. 7c) and matched 1024-sample windows avoiding 4\%--41\% \mbox{long-gap} loss.

\subsection{Generalization Experiments}

\label{h-s5d}

Training the identical architecture and hyperparameters for neutron porosity (NPHI) and photoelectric factor (PEF) confirms modality generality without retuning. Using gamma ray, compressional slowness, caliper and dual resistivity across 13 anchored scenarios (five seeds), \mbox{ST-KF} lowers pooled RMSE by 0.7\%--4.1\% for NPHI (0.0455, 0.0383 and 0.0336 across the three blind cohorts) and by 1.3\%--2.9\% for PEF (1.383, 0.976 and 1.314) against the strongest of Transformer, SAITS, BiGRU and linear interpolation. Linear interpolation edges \mbox{single-point} NPHI gaps, while adding \mbox{whole-well} loss shifts narrow pooled leads to SAITS on \mbox{official-hidden} NPHI and BiGRU on spatially blind PEF.

\begin{table*}[!t]

\markrevisedfloat

\centering\footnotesize\setlength{\tabcolsep}{2pt}

\caption{\rev{\mbox{Zero-shot} RMSE on 149 Taranaki Basin wells (models trained on FORCE 2020 only, mean of five seeds). Markers as in Table II.}}

\label{tbl-tab3}

\input{tables/r1_tbl-tab3.tex}

\end{table*}

\mbox{Zero-shot} transfer to New Zealand's Taranaki Basin reveals a clear decoupling between transferable borehole continuity and \mbox{domain-sensitive} \mbox{cross-property} mappings (Table V, Fig. 6d). Without retraining across distinct lithologies and contractors, \mbox{ST-KF} cuts anchored RMSE by 14\%--68\% below all five baselines, winning 114--147 of 132--147 wells (\(p_{\text{Holm}}<10^{-4}\)). Linear interpolation ranks second on 5-m DTC gaps whereas \mbox{whole-well} \(R^2\) drops to 0.07--0.55 across all models (Transformer edging DTC by 2.0\%, \(p_{\text{Holm}}=0.003\), and \mbox{ST-KF} beating SAITS and BiGRU on RHOB), showing that local anchor continuity transfers across basins whereas \mbox{whole-well} regression is bounded by regional domain shift.

\section{Discussion and Conclusion}

\label{h-s6}

\subsection{Synergistic Mechanism of \mbox{Anchor-Guided} \mbox{State-Space} Fusion}

\label{h-s6a}

\mbox{ST-KF} outperforms unconstrained attention, convolutional and recurrent models because its operators mirror borehole measurement and degradation physics rather than scaling \mbox{black-box} capacity. Field reconstruction faces coupled tool noise and washout or \mbox{cycle-skipping} spikes, boundary offsets from unconstrained amplitude regression and outlier pull or \mbox{high-persistence} stiffness in stationary Gaussian smoothers. Whereas detached wavelet preprocessing \cite{honorio2012well} leaks surviving spikes and unconstrained sequence or imputation networks \cite{gama2025imputation,du2023saits,lin2023missing,meng2024missing} leave \mbox{sub-metre} boundary steps, orthogonalizing these bottlenecks couples multiresolution shrinkage, \mbox{distance-gated} chord residuals and \mbox{variance-decoupled} Rauch--Tung--Striebel smoothing. Multiresolution and firm shrinkage suppress corruption while preserving thin beds, distance gating turns sequence prediction into continuous chord corrections, and conditional tridiagonal smoothing weights priors against clean anchors without manual regime switching. With 0.09\% added parameters and 22.4 ms \mbox{per-well} latency (Table III), this alignment eliminates the 38.7\%--58.8\% \mbox{fusion-removal} and 9.7\%--15.9\% \mbox{gate-removal} \mbox{short-gap} penalties (Table IV), cutting \mbox{blind-cohort} RMSE by 5.0\%--8.3\% for RHOB and 6.4\%--15.6\% for DTC (Table I) while tracking empirical error monotonically via \(\sigma_t\) (Spearman 0.44, 93.3\% and 92.1\% nominal 95\% coverage, Fig. 7a).

\subsection{Paradigm Shift for \mbox{Cross-Basin} \mbox{Well-Log} Reconstruction}

\label{h-s6b}

Beyond accuracy gains, these results shift subsurface log reconstruction from \mbox{basin-specific} amplitude regression toward \mbox{measurement-guided}, learning--inference decoupled \mbox{state-space} fusion. Existing deep models scale recurrent, convolutional, attention or generative backbones \cite{zhang2018lstm,jiang2022gnn,guo2026gan,zhou2025missing,wang2026sonic,meng2024missing} assuming \mbox{cross-property} mappings transfer intact across basins. \mbox{Zero-shot} transfer refutes that assumption and reveals a clear decoupling between local stratigraphic continuity and empirical \mbox{cross-property} inference, as \mbox{ST-KF} widens anchored RMSE gains to 14\%--68\% and linear interpolation ranks second on short sonic gaps whereas \mbox{whole-well} \(R^2\) falls to 0.07--0.55 across all models (Table V). Borehole continuity reflects \mbox{depth-adjacent} stratigraphic persistence and transfers robustly across evaluated basins, whereas unconstrained \mbox{cross-property} mappings shift with regional geology, mineralogy and acquisition conditions. Causal ablation further shows that \mbox{pre-cleaning} training chords induces shortcut learning \cite{geirhos2020shortcut} (\(c=0.28\)--0.39), whereas training on spiky chords and despiking only at inference preserves boundary \mbox{cross-property} sensitivity (\(c\approx0\), Fig. 7d). Density and sonic logs serve as representative targets because they define acoustic impedance, tie wells to seismic data and are among the logs most often degraded by borehole conditions. Training the identical architecture per target extends directly to neutron porosity and photoelectric logs (Section \mbox{V-D}), enabling one frozen estimator per target pair to rehabilitate legacy archives for 3D seismic inversion, geomechanics and frontier hydrocarbon or CO\(_2\) storage characterization.

\subsection{Limitations and Future Directions}

\label{h-s6c}

Several scope conditions constrain the present study and point toward three future directions. First, on isolated 0.15--0.46-m point gaps \mbox{two-point} interpolation is already \mbox{near-optimal}, whereas in completely unlogged wells (\(M_t\equiv0\)) the smoother receives no observations and reduces to the \mbox{cross-property} prior, where \mbox{inter-basin} \mbox{rock-physics} shifts and extreme unanchored noise favor \mbox{multi-basin} pretraining and explicit \mbox{rock-physics} constraints \cite{zhao2023rock,chen2020physics}. Second, replacing \mbox{run-length} firm thresholds and \mbox{prior-only} variance \(\sigma_t^2\) with \mbox{heavy-tailed} Student's \(t\) or \mbox{scale-mixture} likelihoods \cite{zhu2021novel,jia2020novel} estimated jointly with tridiagonal smoothing can calibrate posterior intervals across all regimes. Third, extending validation beyond public wireline suites in two basins to \mbox{logging-while-drilling}, \mbox{high-angle} or horizontal wells and azimuthal anisotropy or image logs \cite{zhao2025anisotropy} will broaden unconventional and geoenergy deployment.

\subsection{Conclusion}

\label{h-s6d}

To overcome spike distortion, \mbox{short-gap} boundary offsets and \mbox{cross-basin} degradation in wireline log reconstruction, this study proposed \mbox{boundary-gated} \mbox{soft-thresholding} Kalman fusion (\mbox{ST-KF}), unifying multiresolution wavelet shrinkage, \mbox{distance-gated} residual learning and \mbox{closed-form} Rauch--Tung--Striebel smoothing under learning--inference decoupling. With 0.09\% parameter overhead and 22.4 ms \mbox{per-well} latency, \mbox{ST-KF} ranks first across 0.15--100-m anchored gaps on blind North Sea cohorts, cutting pooled RMSE by 5.0\%--8.3\% (RHOB) and 6.4\%--15.6\% (DTC) against nine baselines and by 0.7\%--4.1\% (NPHI) and 1.3\%--2.9\% (PEF) against four baselines, while eliminating up to 58.8\% \mbox{fusion-removal} penalties and cutting \mbox{zero-shot} Taranaki anchored errors by 14\%--68\%.

Methodologically, this work shows that boundary mismatch rather than backbone capacity limits deep reconstructors on short gaps, and that local borehole continuity transfers across basins whereas empirical \mbox{cross-property} mappings shift with regional geology. Reconciling \mbox{corruption-resilient} residual priors with \mbox{closed-form} \mbox{state-space} smoothing establishes a lightweight, transferable paradigm for \mbox{legacy-log} quality control, well--seismic calibration and subsurface reservoir characterization.

\section*{Data and Code Availability}

The FORCE 2020 data are available from the FORCE consortium release \cite{bormann2020force} and the Taranaki Basin data from the \mbox{well-log} imputation benchmark \cite{gama2025imputation}. No datasets, well-split files, masks, per-sample predictions, model checkpoints, or data attachments are distributed with this preprint.

\appendices

\section{Training and Hyperparameter Selection}

Training windows are drawn from the 66 training wells. Twenty percent of windows lose the whole target, and the rest receive one to four gaps with \mbox{log-uniform} lengths of 1--700 samples, so a single model covers all regimes. In 50\% of windows, 1\% of the observed target samples per log start a one- to \mbox{three-sample} spike of \(\pm3\)--\(6\) standardized units, while the loss is computed against the clean truth. The loss is the masked mean squared error plus \(0.1\) times the Gaussian negative \mbox{log-likelihood} of the backbone prediction under \(\sigma_t^2\). Optimization uses AdamW (learning rate \(5\times10^{-4}\), weight decay \(10^{-2}\)), a \mbox{one-cycle} schedule, an exponential moving average of weights (0.999), bfloat16 precision, at most 100 epochs and early stopping on validation error with patience 20.

The learning rate of each neural baseline was chosen from \(\{10^{-3},5\times10^{-4}\}\) on the validation wells, and SAITS was run with PyPOTS version 1.5. All models were implemented in Python 3.10 with PyTorch 2.13, PyPOTS 1.5, LightGBM 4.7 and SciPy 1.15, and were trained and evaluated on NVIDIA RTX 5080 graphics processors. Complete architectural dimensions, implementation sources and hyperparameter settings for all baselines are listed in Supplementary Section S7 (Table S18).

The \mbox{ST-KF} hyperparameters were chosen in two \mbox{validation-only} stages. Nine architecture and fusion variants (deeper or wider backbones, window lengths of 512 and 2048, gate floors, wider \(R\) ranges and auxiliary losses) were first screened with five seeds, and none improved on the configuration above. The inference thresholds of the \mbox{front-end} were then set on the validation wells. Sensitivity results are reported in Section \mbox{V-C}. No test well was used for any choice.

\end{revblock}

\IEEEtriggeratref{85}

\end{document}

%% file: tables/r1_tbl-tab1.tex
\begin{tabular}{LCCCCCCCCCCCC}
\toprule
 & \multicolumn{4}{C}{\mbox{In-distribution} blind (18 wells)} & \multicolumn{4}{C}{Spatially blind (10 wells)} & \multicolumn{4}{C}{Official hidden (10 wells)} \\
 & \multicolumn{2}{C}{RHOB} & \multicolumn{2}{C}{DTC} & \multicolumn{2}{C}{RHOB} & \multicolumn{2}{C}{DTC} & \multicolumn{2}{C}{RHOB} & \multicolumn{2}{C}{DTC} \\
Method & RMSE & \(R^2\) & RMSE & \(R^2\) & RMSE & \(R^2\) & RMSE & \(R^2\) & RMSE & \(R^2\) & RMSE & \(R^2\) \\
\midrule
\mbox{ST-KF} & \textbf{0.0663} & \textbf{0.885} & \textbf{5.374} & \textbf{0.951} & \textbf{0.0585} & \textbf{0.932} & \textbf{4.464} & \textbf{0.974} & \textbf{0.0508} & \textbf{0.930} & \textbf{4.375} & \textbf{0.977} \\
Transformer & 0.0706 & 0.870 & 5.741 & 0.944 & 0.0617 & 0.925 & 5.114 & 0.966 & 0.0554 & 0.917 & 5.345 & 0.966 \\
SAITS & 0.0704 & 0.871 & 5.835 & 0.942 & 0.0640 & 0.919 & 5.249 & 0.965 & 0.0557 & 0.916 & 5.181 & 0.968 \\
BiGRU & 0.0753 & 0.852 & 6.090 & 0.937 & 0.0675 & 0.910 & 5.545 & 0.961 & 0.0594 & 0.904 & 6.228 & 0.954 \\
\mbox{U-Net} & 0.0752 & 0.853 & 6.093 & 0.937 & 0.0668 & 0.912 & 5.531 & 0.961 & 0.0599 & 0.903 & 6.260 & 0.953 \\
BiLSTM & 0.0824 & 0.823 & 6.611 & 0.926 & 0.0725 & 0.896 & 6.069 & 0.953 & 0.0680 & 0.874 & 8.034 & 0.923 \\
Linear interp. & 0.1096 & 0.687 & 10.345 & 0.818 & 0.1040 & 0.787 & 10.025 & 0.871 & 0.1032 & 0.710 & 10.402 & 0.871 \\
LightGBM & 0.1199 & 0.626 & 10.728 & 0.804 & 0.1113 & 0.756 & 10.944 & 0.846 & 0.1139 & 0.647 & 11.509 & 0.842 \\
Autoformer & 0.1755 & 0.198 & 21.409 & 0.219 & 0.2039 & 0.180 & 24.280 & 0.243 & 0.1685 & 0.228 & 25.367 & 0.232 \\
\mbox{ST-KF-Autoformer}‡ & 0.1579 & 0.350 & 18.201 & 0.436 & 0.1718 & 0.418 & 20.171 & 0.478 & 0.1465 & 0.416 & 21.768 & 0.434 \\
\bottomrule
\end{tabular}

%% file: tables/r1_tbl-tab2.tex
\begin{tabular}{LCCCCCCCCCCCC}
\toprule
 & \multicolumn{6}{C}{RHOB (g/cm\(^3\))} & \multicolumn{6}{C}{DTC (\(\mu\)s/ft)} \\
Scenario & \mbox{ST-KF} & Transf. & SAITS & BiGRU & \mbox{U-Net} & Linear & \mbox{ST-KF} & Transf. & SAITS & BiGRU & \mbox{U-Net} & Linear \\
\midrule
Point (1--3 samples) & 0.0216 & 0.0354* & 0.0418* & 0.0274 & 0.0370* & \textbf{0.0214} & \textbf{1.397} & 3.153* & 3.522* & 2.722* & 3.971* & 1.438 \\
1 m & \textbf{0.0326} & 0.0401* & 0.0416* & 0.0358 & 0.0437* & 0.0348 & \textbf{2.451} & 3.797* & 3.990* & 3.546* & 4.759* & 2.543 \\
3 m & \textbf{0.0352} & 0.0411* & 0.0417* & 0.0407* & 0.0447* & 0.0487* & \textbf{3.104} & 4.099* & 4.019* & 4.002* & 4.856* & 4.580* \\
5 m & \textbf{0.0453} & 0.0509* & 0.0503* & 0.0527* & 0.0595* & 0.0734* & \textbf{3.417} & 3.997* & 4.205* & 4.310* & 4.991* & 4.704* \\
10 m & \textbf{0.0440} & 0.0488* & 0.0482* & 0.0503* & 0.0527* & 0.0648* & \textbf{3.757} & 4.306 & 4.310* & 4.826* & 5.254* & 5.579* \\
20 m & \textbf{0.0452} & 0.0491* & 0.0485* & 0.0533* & 0.0534* & 0.0669* & \textbf{3.914} & 4.435* & 4.396* & 4.939* & 5.293* & 5.923* \\
50 m & \textbf{0.0483} & 0.0523* & 0.0519* & 0.0568* & 0.0548* & 0.0883* & \textbf{4.235} & 4.476* & 4.521* & 5.000* & 5.037* & 8.689* \\
100 m & \textbf{0.0538} & 0.0585* & 0.0578 & 0.0663* & 0.0615* & 0.0983* & \textbf{4.900} & 5.171 & 5.201 & 5.761 & 5.695 & 9.646* \\
Natural pattern & \textbf{0.0447} & 0.0502* & 0.0495* & 0.0523* & 0.0551* & 0.0755* & \textbf{3.591} & 4.494* & 4.492* & 4.838* & 5.544* & 5.481* \\
\mbox{One-sided} 10 m & \textbf{0.0669} & 0.0671 & 0.0726 & 0.0698 & 0.0733 & 0.1180 & \textbf{4.197} & 5.008 & 5.131 & 6.005 & 6.151 & 8.146* \\
\mbox{One-sided} 50 m & \textbf{0.0648} & 0.0682 & 0.0711 & 0.0742 & 0.0721 & 0.1276* & \textbf{4.767} & 5.820 & 5.622 & 7.393* & 6.925* & 10.538* \\
Whole well & 0.0922 & 0.0913 & \textbf{0.0911} & 0.0969 & 0.0954 & -- & \textbf{8.755} & 8.919 & 8.927 & 9.442 & 9.441 & -- \\
5 m, spiky anchors & \textbf{0.0454} & 0.0517* & 0.0503* & 0.0538* & 0.0615* & 0.1387* & \textbf{3.435} & 4.162* & 4.222* & 4.390* & 5.200* & 10.522* \\
20 m, spiky anchors & \textbf{0.0454} & 0.0499* & 0.0485* & 0.0539* & 0.0551* & 0.1327* & \textbf{3.945} & 4.558 & 4.401 & 5.121* & 5.652* & 15.339* \\
20 m, noise 0.4 & \textbf{0.0530} & 0.0846* & 0.0807* & 0.0853* & 0.0691* & -- & \textbf{4.866} & 7.238* & 7.941* & 7.806* & 6.775* & -- \\
Whole well, noise 0.4 & \textbf{0.1025} & 0.1362* & 0.1452* & 0.1294* & 0.1144 & -- & 11.182 & 14.503* & 16.782* & 12.387* & \textbf{11.068} & -- \\
\bottomrule
\end{tabular}

%% file: tables/r1_tbl-tab5.tex
\begin{tabular}{LCCCC}
\toprule
Model & Params & Train (min) & ms/well & DTC RMSE \\
\midrule
\mbox{ST-KF} (frozen) & 597,793 & 13.9 & 22.4¶ & 4.737 \\
\mbox{ST-KF} w/o \mbox{inference-time} modules & 597,793 & 13.9 \(\pm\) 1.0 & 17.8 & 4.771 \\
Transformer & 597,250 & 7.1 \(\pm\) 0.7 & 5.4 & 5.400 \\
SAITS & 540,798 & 24.1 \(\pm\) 5.5 & 24.7 & 5.422 \\
BiGRU & 407,554 & 6.5 \(\pm\) 1.0 & 10.8 & 5.954 \\
BiLSTM & 543,234 & 7.7 \(\pm\) 1.1 & 22.4 & 6.905 \\
\mbox{U-Net} & 1,065,570 & 7.3 \(\pm\) 0.8 & 7.8 & 5.961 \\
\bottomrule
\end{tabular}

%% file: tables/r1_tbl-tab4.tex
\begin{tabular}{LCCCCCCCCCC}
\toprule
 & \multicolumn{2}{C}{\(\le\)1-m gaps} & \multicolumn{2}{C}{3--100-m gaps} & \multicolumn{2}{C}{Whole well} & \multicolumn{2}{C}{Spiky anchors} & \multicolumn{2}{C}{Noisy inputs} \\
Variant & RHOB & DTC & RHOB & DTC & RHOB & DTC & RHOB & DTC & RHOB & DTC \\
\midrule
\multicolumn{11}{L}{(a) \mbox{Inference-time} modules}\\
w/o anchor bridge cleaning & +0.0 & -0.0 & -0.0 & +0.1 & +0.0 & +0.0 & +2.7 & +5.7 & +0.0 & +0.0 \\
w/o anchor cleaning and \mbox{Firm-IRLS} & -0.1 & -0.0 & -0.0 & +0.1 & +0.0 & +0.0 & +3.0 & +7.1 & +0.0 & +0.0 \\
w/o all \mbox{inference-time} modules & -0.1 & -0.2 & -0.2 & -0.0 & -0.1 & +0.0 & +3.0 & +7.1 & +26.4 & +16.5 \\
\multicolumn{11}{L}{(b) Trained components}\\
w/o \mbox{distance-gated} bridge & +9.7 & +15.9 & +4.8 & +8.0 & -0.5 & +1.0 & +1.5 & +1.1 & +5.5 & +5.6 \\
coupled \(q_t\) and \mbox{smoothed-residual} outlier test & +11.7 & +31.6 & +3.8 & +4.6 & -0.3 & +1.9 & +1.4 & -0.4 & +2.3 & +2.6 \\
w/o innovation shrinkage (wavelet only) & +7.5 & +12.3 & +4.1 & +7.8 & -0.2 & +1.4 & +1.0 & +0.7 & +4.1 & +3.6 \\
w/o input wavelet shrinkage§ & +11.6 & +32.3 & +3.4 & +4.8 & -0.2 & +1.5 & +0.8 & -0.1 & +15.7 & +10.2 \\
w/o any shrinkage & +6.9 & +12.5 & +3.4 & +7.6 & -0.3 & +1.8 & +0.4 & +0.7 & +16.9 & +12.5 \\
w/o Kalman fusion§ & +38.7 & +58.8 & +4.4 & +9.3 & -1.5 & +0.1 & +1.8 & +2.9 & -- & -- \\
\bottomrule
\end{tabular}

%% file: tables/r1_tbl-tab3.tex
\begin{tabular}{LCCCCCCCC}
\toprule
 & \multicolumn{2}{C}{5-m gaps} & \multicolumn{2}{C}{20-m gaps} & \multicolumn{2}{C}{Natural pattern} & \multicolumn{2}{C}{Whole well} \\
Method & RHOB & DTC & RHOB & DTC & RHOB & DTC & RHOB & DTC \\
\midrule
\mbox{ST-KF} & \textbf{0.0373} & \textbf{3.373} & \textbf{0.0462} & \textbf{4.254} & \textbf{0.0476} & \textbf{4.168} & 0.1314 & 11.255 \\
Transformer & 0.0499* & 4.281* & 0.0594* & 5.075* & 0.0650* & 5.371* & 0.1317 & \textbf{11.031†} \\
SAITS & 0.0450* & 4.301* & 0.0538* & 5.008* & 0.0594* & 5.435* & 0.1360* & 11.386 \\
BiGRU & 0.0538* & 4.541* & 0.0659* & 5.530* & 0.0680* & 5.632* & 0.1352* & 11.507 \\
Linear interp. & 0.0498* & 4.295* & 0.0654* & 5.873* & 0.0680* & 5.356* & -- & -- \\
LightGBM & 0.1167* & 10.102* & 0.1161* & 10.188* & 0.1268* & 10.766* & \textbf{0.1281} & 11.368 \\
\bottomrule
\end{tabular}